\documentclass{article}

\usepackage{microtype}
\usepackage{graphicx}
\usepackage{subfigure}
\usepackage{booktabs} % for professional tables
\usepackage{xspace}

\usepackage{hyperref}

\usepackage[accepted]{icml2024}

\usepackage{amsmath}
\usepackage{amssymb}
\usepackage{mathtools}
\usepackage{amsthm}

\DeclareMathOperator*{\argmax}{arg\,max}
\DeclareMathOperator*{\argmin}{arg\,min}

\usepackage[capitalize,noabbrev]{cleveref}

\theoremstyle{plain}

\theoremstyle{definition}

\theoremstyle{remark}

\usepackage[textsize=tiny]{todonotes}

\makeatletter
\DeclareRobustCommand\onedot{\futurelet\@let@token\@onedot}
\def\@onedot{\ifx\@let@token.\else.\null\fi\xspace}
 
\def\eg{\emph{e.g}\onedot\ } 
\def\ie{\emph{i.e}\onedot\ } 
 
 \def\vs{\emph{vs}\onedot}
\def\wrt{\emph{w.r.t}\onedot\ }

\makeatother

\icmltitlerunning{Few-Shot Unsupervised Implicit Neural Shape Representation Learning with Spatial Adversaries}

\begin{document}

\twocolumn[
\icmltitle{Few-Shot Unsupervised Implicit Neural Shape Representation Learning with Spatial Adversaries}
\icmlsetsymbol{equal}{*}

\begin{icmlauthorlist}
\icmlauthor{Amine Ouasfi}{yyy}
\icmlauthor{Adnane Boukhayma}{yyy}
\end{icmlauthorlist}

\icmlaffiliation{yyy}{Inria, Univ. Rennes, CNRS, IRISA, M2S, France}

\icmlcorrespondingauthor{Adnane  Boukhayma}{adnane.boukhayma@gmail.com}
\vskip 0.3in
]
\printAffiliationsAndNotice{}  

\begin{abstract}
\eg Implicit Neural Representations have gained prominence as a powerful framework for capturing complex data modalities, encompassing a wide range from 3D shapes to images and audio. Within the realm of 3D shape representation, Neural Signed Distance Functions (SDF) have demonstrated remarkable potential in faithfully encoding intricate shape geometry. However, learning SDFs from sparse 3D point clouds in the absence of ground truth supervision remains a very challenging task. While recent methods rely on smoothness priors to regularize the learning, our method introduces a regularization term that leverages adversarial samples around the shape to improve the learned SDFs. Through extensive experiments and evaluations, we illustrate the efficacy of our proposed method, highlighting its capacity to improve SDF learning with respect to baselines and the state-of-the-art using synthetic and real data.
\end{abstract}

\section{Introduction}

Obtaining faith-full and intelligible neural representations of the 3D world from limited and corrupted point clouds is a challenge of paramount importance, that finds applications in countless downstream computer vision and graphics tasks. While many methods rely on data priors learned from large fully labeled datasets, these priors can fail to generalize to unseen shapes especially under very sparse unoriented inputs \cite{NeuralTPS,ouasfi2024robustifying}. Hence, it is important to design learning frameworks that can lead to efficient and robust learning of implicit shape representations under such extreme constraints.   

\begin{figure}[t!]
\centering
\includegraphics[width=0.49\linewidth]{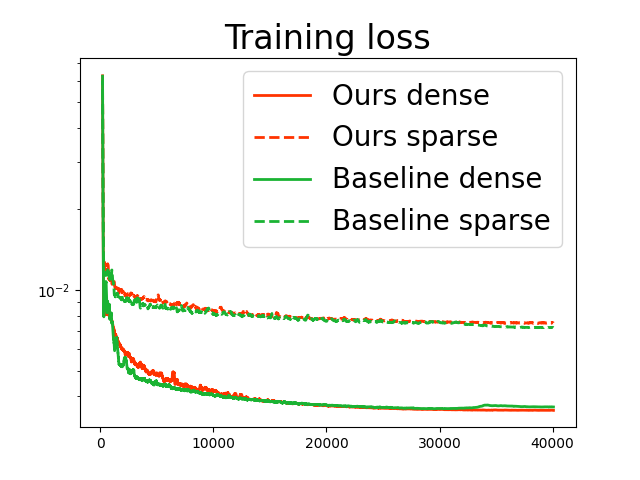}
\includegraphics[width=0.49\linewidth]{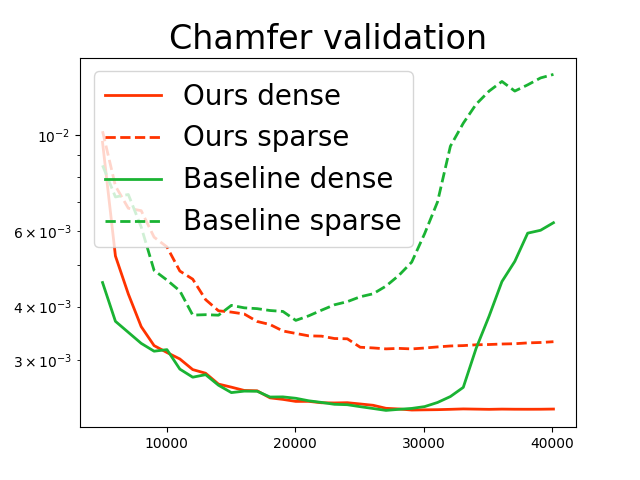}
%\vspace{-10pt}
\caption{While the training loss (left) is decreasing for both our baseline \cite{ma2020neural} and our method, the Chamfer distance of reconstructions \wrt the GT starts increasing quite early on especially in the sparse input point cloud case for the baseline. This undesirable behaviour is remedied by our adversarial query mining. We report here metrics for unit box normalized meshes, using shape Gargoyle of dataset SRB \cite{williams2019deep}.}
\label{fig:motiv}
\end{figure}

In this context, the learning strategy introduced by \cite{ma2020neural} (dubbed NeuralPull) have shown to be one of the most successful ones in learning implicit shapes from point cloud unsupervisedly. However, upon observing the behavior of the training and validation errors of this method under sparse and dense input point clouds (Figure \ref{fig:motiv}), we notice that the validation error starts increasing quite early on in the training in the sparse input case, whilst the training loss keeps on decreasing. This suggests an overfitting problem evidently intensifying in the sparse setting. Qualitatively%(as shown in accompanying comparative videos in the supplementary material)
, this increase in the validation error is usually synonymous to deterioration in the extracted shape with symptoms varying between shape instances, including shape hallucinations, missing shape parts, and shape becoming progressively wavy, bumpy and noisy. In extreme cases, shapes can also break into separate components or clusters around input points. When the input is additionally noisy, these phenomena are further exacerbated. 

Recent work in the field relies on various smoothness priors (\eg  \cite{NeuralTPS,gropp2020implicit,ben2022digs,ouasfi2024unsupervised}) to regularize the implicit shape functions, and hence reduce overfitting. One side of the problem that remains underexplored however is how training data is sampled during learning, and understanding to which extent this sampling could affect performance. This is even the more an important question in our situation. In fact, while standard supervised learning  uses typically data/label sample pairs, fitting implicit representations entails mapping spatial coordinates to labels or pseudo labels, where these spatial queries can be sampled uniformly or normally around the input point cloud. In the case of our baseline NeuralPull, the nearest point could sample to a spatial query is a pseudo-label approximating the unavailable nearest groundtruth surface point in the training.   
Hence, both inherent input point cloud noise and its sparsity represent a \textit{noise} (\ie displacement) on the perfect surface labels. This composite noise  can affect both the SDF function and gradient orientation. In practice, we notice the network first produces a very smooth shape. When it tries to refine it, it tends to overfit to the noise present in the  supervision signal. At this stage, further fitting on easy samples (predominant samples) means overfitting on this noise. The samples that can benefit the implicit representation can be drowned within easy samples.
 
Among literature interested in such a problem, active learning advocates sampling based on informativeness and diversity \cite{huang2010active}. New samples are queried from a pool of unlabeled data given a measure of these criteria. Informative samples are usually defined as samples that can reduce the uncertainty of a statistical model. However, several heuristics coexist as it is impossible to obtain a universal active learning strategy that is effective for any given task \cite{dasgupta2005coarse}. In our setting it is not clear what samples are the most informative for our implicit shape function and how to sample from the uncertainty regions of the implicit field. Recent work on distributionally robust optimization (DRO) \cite{volpi2018generalizing,rahimian2019distributionally} provides a mathematical framework to model uncertainty. In this framework, the loss is minimized  over the worst case distribution in a neighborhood of the observed training data distribution. As a special case, adversarial training \cite{madry2017towards} uses pointwise adversaries rather than adversarial joint perturbations of the entire training set. 

Inspired by this literature, we propose to use adversarial samples to regularize the learning of implicit shape representations from sparse point clouds. We build on SDF projection minimization error loss in training (See figure \ref{fig:overview}). Typically query points are pre-sampled around the input point cloud to train such a method. We augment these queries with adversarial samples during training. To ensure the diversity of these additional samples, we generate them in the vicinity of original queries within locally adapted radii. These radii modulate the adversarial samples density in concordance with the input point cloud density, thus allowing us to adapt to the local specificities of the input during the neural fitting. Our adversarial training strategy, focuses on samples that can still benefit the network, which prevents the aforementioned overfitting while refining the implicit representation. 

To test our idea, we devise experiments on real and synthetic reconstruction benchmarks, including objects, articulated shapes and large scenes. Our method outperforms the baseline as well as the most related competition both quantitatively and qualitatively. We notice that our adversarial loss helps our model most in places where shape prediction is the hardest and most ambiguous, such as fine and detailed structures and body extremities. Experiments on a dense reconstruction setting show that our method can be useful in this setup as well. Finally, thanks to our method, and as illustrated in Figure \ref{fig:motiv}, validation stabilizes and plateaus at convergence unlike our baseline, which makes it easier for us to decide the evaluation model epoch, given that evaluation measurements are normally unavailable in unsupervised settings.

\section{Related work}

Classical shape modelling from point cloud includes combinatorical methods where the shape is defined through an input point cloud based space partitioning, through \eg alpha shapes \cite{bernardini1999ball} Voronoi diagrams \cite{amenta2001power} or triangulation \cite{cazals2006delaunay,liu2020meshing,rakotosaona2021differentiable}. Differently, the input samples can be used to define an implicit function whose zero level set represents the target shape, using global smoothing priors \cite{williams2022neural,lin2022surface,williams2021neural} \eg radial basis function \cite{carr2001reconstruction} and Gaussian kernel fitting \cite{scholkopf2004kernel}, local smoothing priors such as moving least squares \cite{mercier2022moving,guennebaud2007algebraic,kolluri2008provably,liu2021deep}, or by solving a boundary conditioned Poisson equation \cite{kazhdan2013screened}. The recent literature proposes to parameterise these implicit functions with deep neural networks and learn their parameters with gradient descent, either in a supervised 
(\eg  
\cite{ouasfi2024Mixing, ouasfi2022few,boulch2022poco,peng2020convolutional,lionar2021dynamic,peng2021shape}) 
or unsupervised manner. These implicit neural representations \cite{mescheder2019occupancy,park2019deepsdf} overcome many of the limitations of explicit ones
(\eg meshes \cite{wang2018pixel2mesh,kato2018neural,jena2022neural} and point clouds \cite{fan2017point,aliev2020neural,kerbl20233d})
in modelling shape, radiance and light fields (\eg \cite{mildenhall2020nerf,yariv2021volume,wang2021neus,jain2021dreamfields,chan2022efficient,li2023learning,li2023regularizing,jena2024geo,younes2024sparsecraft}), as they allow to represent functions with arbitrary topologies at virtually infinite resolution. 

We are interested in unsupervised implicit neural shape learning. In this scenario, an MLP  is typically fitted to the input point cloud without extra priors or information. Regularizations can compensate for the lack of supervision. For instance, \cite{gropp2020implicit} introduced a spatial gradient constraint based on the Eikonal equation. \cite{ben2022digs} introduces a spatial divergence constraint. \cite{liu2022learning} propose a Lipschitz regularization on the network. 
\cite{ma2020neural,ouasfi2024unsupervised} expresses the nearest point on the surface as a function of the neural signed distance and its gradient. \cite{peng2021shape} proposed a differentiable Poisson solving layer that converts predicted normals into an indicator function grid efficiently.
\cite{koneputugodage2023octree} guides the implicit field learning with an Octree based labelling. \cite{boulch2021needrop} predicts occupancy fields by learning whether a dropped needle goes across the surface or not. \cite{NeuralTPS} learns a surface parametrization leveraged to provide additional coarse surface supervision to the shape network. Most methods can benefit from normals if available. \cite{atzmon2020sal} proposed to supervise the gradient of the implicit function with normals, while \cite{williams2021neural} uses the inductive bias of kernel ridge regression. In the absence of normals and learning-based priors, and under input scarcity, most methods still display considerable failures. Differently from all existing work, we explore the use of adversarial samples in training implicit neural shape representations. 

In the adversarial training literature,  a trade-off between accuracy and robustness has been observed empirically in different datasets \cite{raghunathan2019adversarial,madry2018towards}. This has led prior work to claim that this tradeoff may be inevitable for many classification tasks \cite{tsipras2018robustness,zhang2019theoretically}. However, several recent papers challenged this claim. \cite{yang2020closer} showed theoretically that this tradeoff is not inherent. It is rather a consequence of current robustness methods. These findings are corroborated empirically in recent work \cite{stutz2019disentangling,xie2020adversarial,herrmann2021pyramid}. Our baseline relies on a pseudo-labeling strategy that introduces noise as the input gets sparser. Our method robustifies the learning against this noise, providing regularization and additional informative samples. The regularization helps prevent overfitting and enhances generalization, \ie ensuring the loss behavior on the “training” query points is generalized in the 3D space, while informative samples aid in refining the shape function during advanced training stages. 
\section{Method}

\begin{figure}[t!]
\centering
\includegraphics[width=1.0\linewidth]{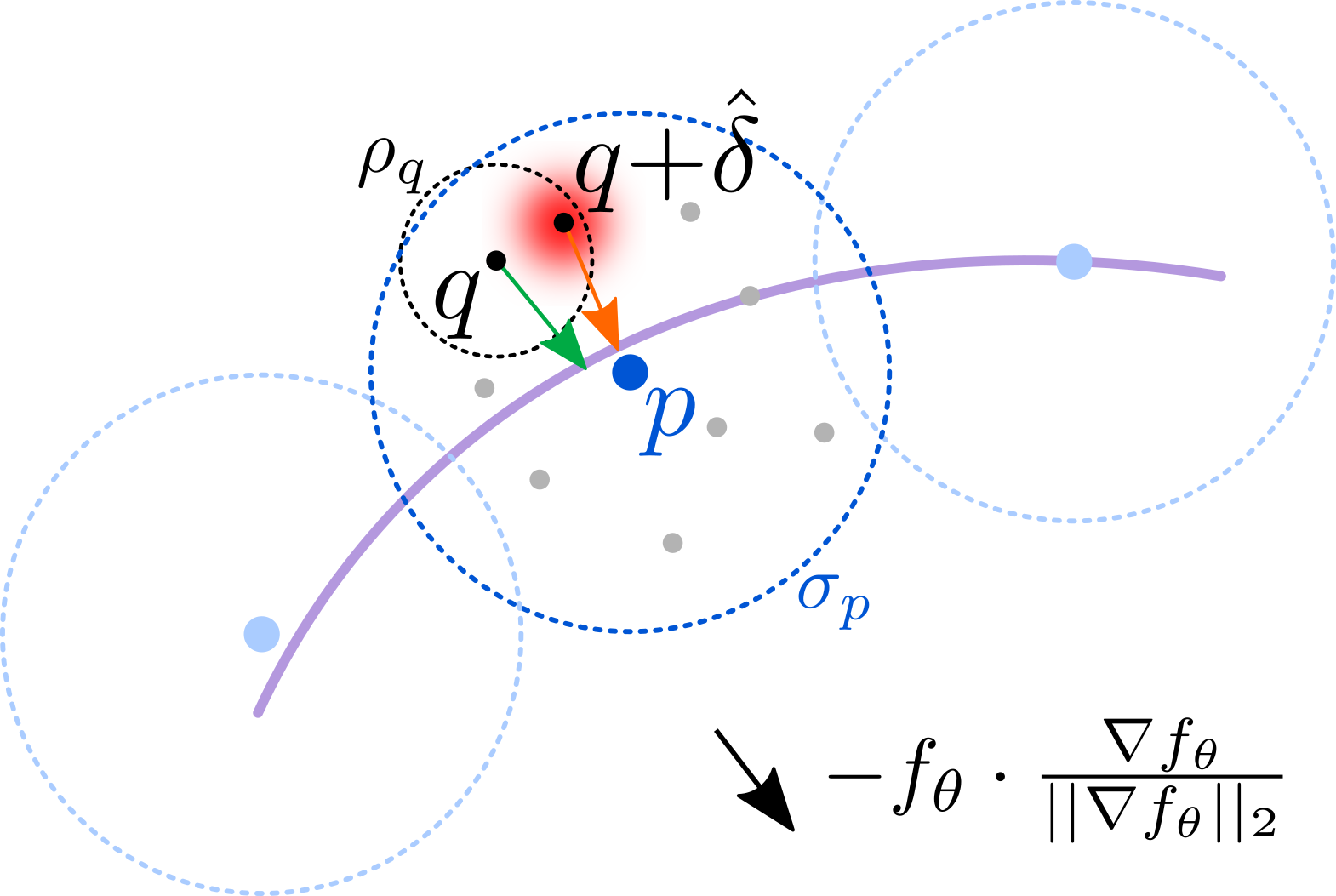}
%\vspace{-15pt}
\caption{We learn an implicit shape representation $f_{\theta}$ from a point cloud (blue points) by minimizing the error between projection (through $f_{\theta}$) of spatial queries $q$ (gray points) onto the level set of the field (purple) and the nearest input point $p$. We introduce adversarial queries $q+\hat{\delta}$ to the optimization. They are defined as samples maximizing the loss in the vicinity of original queries.}
\label{fig:overview}
\end{figure}

Given a noisy, sparse unoriented point cloud $\mathbf{P}\subset{\mathbb R}^{3\times N_p}$, our objective is to obtain a corresponding 3D shape reconstruction, \ie the shape surface  $\mathcal{S}$ that best explains the observation $\mathbf{P}$. In other terms, the input point cloud elements should approximate noised samples from $\mathcal{S}$. 

In order to achieve this goal, we learn a shape function $f$ parameterised with an MLP $f_\theta$. The function represents the implicit signed distance field relative to the target shape $\mathcal{S}$. That is, for a query euclidean space location $q \in {\mathbb R}^{3}$, $f(q):=s \cdot \min_{v\in\mathcal{S}}||v-q||_2$, where $s:=1$ if $q$ is inside $\mathcal{S}$, and $s:=-1$ otherwise. The inferred shape $\hat{\mathcal{S}}$ can be obtained as the zero level set of the SDF (signed distance function) $f_\theta$ at convergence:
\begin{equation}
\hat{\mathcal{S}} = \{ q\in\mathbb{R}^3 \mid f_\theta(q) = 0\}.
\end{equation} 
Practically, an explicit triangle mesh for $\hat{\mathcal{S}}$ can be obtained through the Marching Cubes algorithm \cite{lorensen1987marching}, while querying neural network $f_\theta$. We note also that $\hat{\mathcal{S}}$ can be rendered through ray marching \cite{hart1996sphere} through the SDF field inferred by $f_\theta$. 

\subsection{Background: Learning an SDF by pulling queries onto the surface.}

Several state-of-the-art reconstruction from point cloud methods (\eg \cite{NeuralTPS,ma2022surface,ma2022reconstructing,LPI,ma2020neural}), including the state-of-the-art unsupervised reconstruction from sparse point cloud method \cite{NeuralTPS}, build on the neural SDF training procedure introduced in \cite{ma2020neural} named NeuralPull. 
The latter is inspired by the observation that the distance field guided projection operator $q \mapsto q - f(q) \cdot \nabla f(q)$ (\cite{chibane2020neural,perry2001kizamu,wolter1993cut,zhao2021learning}) yields the nearest surface point when applied near the surface, where $\nabla f$ is the spatial gradient of $f$.   

In practice, query points $q \in Q$  are sampled around the input point cloud $\mathbf{P}$, specifically from normal distributions centered at input samples $\{p\}$, with locally defined standard deviations $\{\sigma_p\}$:
\begin{equation}
Q := \bigcup_{p\in \mathbf{P}}\{q \sim \mathcal{N}(p,\sigma_p \mathbf{I}_3)\},
\label{equ:sample}
\end{equation} 
where $\sigma_p$ is chosen as the maximum euclidean distance to the $K$ nearest  points to $p$ in $\mathbf{P}$. For each query $q$, the nearest point $p$ in $\mathbf{P}$ is computed subsequently, and the following objective is optimized in \cite{ma2020neural} yielding a neural signed distance field $f_\theta$ whose zero level set concurs with the samples in $\mathbf{P}$:
\begin{equation}
\begin{aligned}
&\mathcal{L}(\theta,q) = ||q - f_\theta(q) \cdot \frac{\nabla f_\theta(q)}{||\nabla f_\theta(q)||_2} - p||_2^2, \\
&\text{where} \quad p = \argmin_{t\in\mathbf{P}}||t-q||_2.
\label{equ:np}
\end{aligned} 
\end{equation}

In the case of \cite{ma2020neural}, the training procedure is empirical risk minimization (ERM), namely:
\begin{equation}
\min_\theta \underset{q\sim Q}{\mathbb{E}} \mathcal{L}(\theta,q),
\label{eq:erm}
\end{equation} 
where $Q$ constitutes the empirical distribution over the training samples.  
\subsection{Local adversarial queries for the few shot setting}

As introduced in the first section, input point cloud noise and sparsity are akin to noisy labels for query points $q$. Hence, training through standard ERM under sparse input point clouds $\mathbf{P}$ leads to an overfitting on this noise (Figure \ref{fig:motiv}), \ie useful information carried out by query samples decreasing during training thus leading to poor convergence. Hence, differently from existing work in the field of learning based reconstruction from point cloud, we propose to focus on the manner in which query points $q$ are sampled at training, as we hypothesise that there could be a different sampling strategy from the one proposed in NeuralPull (\ie Equation \ref{equ:sample}) that can lead to better results. This hypothesis is backed by literature showing that hard sample mining can lead to improved generalization and reduced over-fitting \cite{xie2020adversarial,chawla2010data,fernandez2014we,krawczyk2016learning,shrivastava2016training}. Intuitively, exposing the network to the worst cases in training is likely to make it more robust and less specialized on the more common easy queries. 

We explore a different procedure from standard ERM (Equation \ref{eq:erm}). Ideally, we wish to optimize $\theta$ under the worst distribution $Q'$ of query points $\{q\}$ in terms of our objective function, meaning: 
\begin{equation}
\min_\theta \max_{Q'} \underset{q\sim Q'}{\mathbb{E}} \mathcal{L}(\theta,q).
\end{equation}
Such a training procedure is akin to a distributionally robust optimization \cite{Sagawa*2020Distributionally,rahimian2019distributionally} which is hard to achieve in essence. It was shown however that a special more attainable case of the latter consists in harnessing hard samples locally \cite{staib2017distributionally}, that is looking for hard samples in the vicinity of the original empirical samples:
\begin{equation}
\min_\theta \underset{q\sim Q}{\mathbb{E}} \max_{\delta,\, ||\delta||_2\leq\rho} \mathcal{L}(\theta,q+\delta).
\end{equation}
%\adnane[]{say something like: notice that in our case, in order to use the same target point $p$ in the loss $mathcal{L}$ for query $q+\delta$ ($p$ being the nearest point to $q$) in $\mathbf{P}$), $\delta$ needs to remain}

Let us consider optimum perturbations $\hat{\delta}$. Using a first order Taylor expansion on loss $\mathcal{L}(\theta,q+\delta)$, we can write:
\begin{equation}
\begin{aligned}
\hat{\delta} & = \argmax_{|| \delta||_2 \leq\rho}\mathcal{L}(\theta,q+\delta),\\
& \approx \argmax_{|| \delta||_2 \leq\rho} \mathcal{L}(\theta,q) + \delta^\top \nabla_q\mathcal{L}(\theta,q),\\
& \approx \argmax_{|| \delta||_2 \leq\rho}\delta^\top \nabla_q \mathcal{L}( \theta,q).
\end{aligned}
\end{equation}

Leveraging this approximation, we can obtain the optimum value $\hat{\delta}$ as the solution to a classical dual norm problem, by using the equality case of Cauchy–Schwarz inequality applied to the scalar product $\delta^\top \nabla_q \mathcal{L}( \theta,q)$ for $\delta$ in a closed ball of radius $\rho$ and center $0$:
\begin{equation}
\hat{\delta} = \rho  \,\frac{\nabla_q\mathcal{L}(\theta,q)}{||\nabla_q\mathcal{L}(\theta,q)||_2},
\label{equ:delta}
\end{equation}
where gradients $\nabla_q\mathcal{L}$ can be computed efficiently through automatic differentiation in a deep-learning framework (\eg PyTorch \cite{paszke2019pytorch}). 

\begin{algorithm}[h!]
\small
\begin{algorithmic}
\REQUIRE Point cloud $\textbf{P}$, learning rate $\alpha$, number of iterations $N_{\text{it}}$, batch size $N_b$.
\ENSURE Optimal parameters ${\theta}^*$.
\STATE Compute local st. devs. $\{\sigma_p\}$ ($\sigma_p=\max_{t\in K\text{nn}(p,\mathbf{P})}||t-p||_2$).
\STATE $Q \leftarrow$ sample($\textbf{P}$,$\{\sigma_p\}$) (Equ. \ref{equ:sample})
\STATE Compute nearest points in $\textbf{P}$ for all samples in $Q$.
\STATE Compute local radii $\{\rho_q\}$ $\left(\rho_q=\sigma_p\times 10^{-2},\; p := \text{nn}(q,\mathbf{P})\right)$.
\STATE Initialize $\lambda_1 = \lambda_2 = 1$. 
\FOR{$N_{\text{it}}$ times}
%\textbf{for} $N_{\text{it}}$ times \textbf{do}  
\STATE Sample $N_b$ query points $\{q, q\sim Q\}$.
\STATE Compute losses $\{\mathcal{L}(\theta,q)\}$. (Equ. \ref{equ:np})
\STATE Compute loss gradients $\{\nabla_q\mathcal{L}(\theta,q)\}$ with autodiff.
\STATE Compute 3D offsets $\{\hat{\delta}\}$. (Equ. \ref{equ:delta}, using radii $\{\rho_q\}$)
\STATE Compute adversarial losses $\{\mathcal{L}(\theta,q+\hat{\delta})\}$. (Equ. \ref{equ:np})
\STATE Compute combined losses $\{\mathfrak{L}(\theta,q)\}$. (Equ. \ref{equ:final})
\STATE $(\theta,\lambda_1,\lambda_2) \leftarrow (\theta ,\lambda_1,\lambda_2) - \alpha \nabla_{\theta,\lambda_1,\lambda_2} \Sigma_q \mathfrak{L}(\theta ,q)$
%\textbf{end for}
%\EndFor
\ENDFOR
\end{algorithmic}
\caption{\small The training procedure of our method.}
\label{alg:train}
\end{algorithm}

We found empirically that using local radii $\{\rho_q\}$ in our context  improves over using a single global radius $\rho$ and we provide and ablation later on of this design choice. We recall that each query point $q$ has a nearest counterpart $p$ in $\mathbf{P}$. As we want our adversarial sample $q+\hat{\delta}$ to still remain relatively close to $p$, we define $\{\rho_q\}$ as a fraction of local standard deviation $\sigma_p$ of the nearest point $p$ (\eg $\rho_q = \sigma_p \times 10^{-2}$ \label{equ:rad}). Linking $\rho_q$ to $\sigma_p$ also allows us to adjust the local radii to the local input point cloud density. 
We remind that $\sigma_p$ was used for sampling query points $q$ around input point $p$, as explained in the previous background section.

To ensure the stability of our learning, we train our neural network by backpropagating a hybrid loss combining the original objective and the adversarial one, using the strategy in \cite{liebel2018auxiliary} for multi-task learning:
\begin{equation}
\begin{aligned}
     \mathfrak{L}(\theta,q) = &\frac{1}{2\lambda_1}\mathcal{L}(\theta,q) + \frac{1}{2\lambda_2}\mathcal{L}(\theta,q+\hat{\delta})\\ 
     &+ \ln{(1+\lambda_1)} + \ln{(1+\lambda_2)},
\end{aligned}
\label{equ:final}    
\end{equation}
where $\lambda_1$ and  $\lambda_2$ are learnable weights. A summary of our training procedure is shown in Algorithm \ref{alg:train}. A visual illustration of our training can be seen in Figure \ref{fig:overview}.

%\begin{algorithm}[tb]
%   \caption{Bubble Sort}
%   \label{alg:example}
%\begin{algorithmic}
%   \STATE {\bfseries Input:} data $x_i$, size $m$
%   \REPEAT
%   \STATE Initialize $noChange = true$.
%   \FOR{$i=1$ {\bfseries to} $m-1$}
%   \IF{$x_i > x_{i+1}$}
%   \STATE Swap $x_i$ and $x_{i+1}$
%   \STATE $noChange = false$
%   \ENDIF
%   \ENDFOR
%   \UNTIL{$noChange$ is $true$}
%\end{algorithmic}
%\end{algorithm}

\section{Results}

To evaluate our method, we assess our ability to learn implicit shape representations given sparse and noisy point clouds.  We use datasets from standard reconstruction benchmarks.  
These datasets highlight a variety of challenges of fitting coordinate based MLPs to sparse data as well as reconstruction more broadly. Following the literature, we evaluate our method by measuring the accuracy of 3D explicit shape models extracted after convergence from our MLPs. 
We compare quantitatively and qualitatively to the the state-of-the-art in our problem setting, \ie unsupervised reconstruction from unoriented point cloud, including methods designed for generic point cloud densities and methods dedicated to the sparse setting. For the former, we compare to  fully implicit  deep learning methods such as NP \cite{ma2020neural}, SAP \cite{peng2021shape}, DIGS \cite{ben2022digs}, in addition to hybrid methods combining implicit and grid based representations such as  OG-INR \cite{koneputugodage2023octree} and GP (GridPull) \cite{chen2023gridpull}. When it comes to methods dedicated to the sparse setting we compare to NTPS \cite{NeuralTPS} which is the closest method to ours as it focuses specifically on the sparse input case. We additionally compare to NDrop \cite{boulch2021needrop}. We show results for NSpline \cite{williams2021neural} even though it requires normals. We also compare to classical Poisson Reconstruction SPSR \cite{kazhdan2013screened}. We note also that comparisons to NP (NeuralPull, our baseline) also serves additionally as an ablation of our adversarial loss through out our experiments. For comprehensive evaluation, we also include comparisons to supervised methods including  state-of-the-art feed-forward generalizable methods, namely POCO \cite{boulch2022poco}, CONet \cite{peng2020convolutional} and NKSR \cite{Huang_2023_CVPR}, alongside the finetuning method SAC \cite{tang2021sa} and the prior-based optimization method dedicated to sparse inputs On-Surf \cite{ma2022reconstructing}.  Unless stated differently, we use the publicly available official implementations of existing methods.
For sparse inputs, we experimented with point clouds of size $N_p = 1024$. %While our main focus here is learning SDFs from sparse, noisy and unoriented point clouds, we found that addressing all these challenges under extremely sparse inputs (e.g. 300) leads in many cases to reconstructions that are not very meaningful or useful, and which are also hard to assess properly with standard metrics. Implementation details are provided in the appendix.

\subsection{Metrics}

Following seminal work, we evaluate our method and the competition \wrt the ground truth using standard metrics for the 3D reconstruction task. Namely, the L1 \textbf{Chamfer Distance
CD$_1$} ($\times10^{2}$), L2 \textbf{Chamfer Distance
CD$_2$} ($\times10^{2}$), the euclidean distance based \textbf{F-Score (FS)} when ground truth points are available, and finally \textbf{Normal Consistency (NC)} when ground truth normals are available. We detail the expressions of these metrics in the appendix.

\subsection{Datasets and input definitions}

\textbf{ShapeNet} \cite{shapenet} consists of various instances of 13 different synthetic 3D object classes. We follow the train/test splits defined in \cite{williams2021neural}. We generate noisy input point clouds by sampling $1024$ points from the meshes and adding Gaussian noise of standard deviation $0.005$ following the literature (\eg \cite{boulch2022poco,peng2020convolutional}). For brevity we show results on classes Tables, Chairs and Lamps. 

\textbf{Faust} \cite{Bogo:CVPR:2014} consists of real scans of 10 human body identities in 10 different poses. We sample sets of $1024$ points from the scans as inputs. 

\textbf{3D Scene} \cite{zhou2013dense} contains large scale complex real world scenes obtained with a handheld commodity range sensor. We follow \cite{NeuralTPS,jiang2020local,ma2020neural} and sample our input point clouds with a sparse density of $100$ per m$^2$, and we report performance similarly for scenes Burghers, Copyroom, Lounge, Stonewall and Totempole.    

\textbf{Surface Reconstruction Benchmark (SRB)} \cite{williams2019deep} consists of five object scans, each with different challenges such as complex topology, high level of detail, missing data and varying feature scales. We sample $1024$ points from the scans for the sparse input experiment, and we also experiment using the dense inputs.

\subsection{Implementation details}

%For sparse inputs, we experimented with point clouds of size $N_p = 1024$. While our main focus here is learning SDFs from sparse, noisy and unoriented point clouds, we found that addressing all these challenges under extremely sparse inputs (e.g. 300) leads in many cases to reconstructions that are not very meaningful or useful, and which are also hard to assess properly with standard metrics.
Our MLP ($f_\theta$) follows the architecture in NP \cite{ma2020neural}. We train for $N_{it}=40000$  iterations using the Adam optimizer. %We use learning rate $\alpha=10^$, 
We use batches of size $N_b=5000$. Following NP, we set $K=51$ for estimating local standard deviations $\sigma_p$. We train on a NVIDIA RTX A6000 GPU. Our method takes $8$ minutes in average  to converge for a $1024$ sized input point cloud. In the interest of practicality and fairness in our comparisons, we decide the evaluation epoch for all the methods for which we generated results (including our main baseline) in the same  way: we chose the best epoch for each method in terms of chamfer distance between the reconstruction and the input point cloud.

\begin{figure}[t!]
\centering
\includegraphics[width=\linewidth]{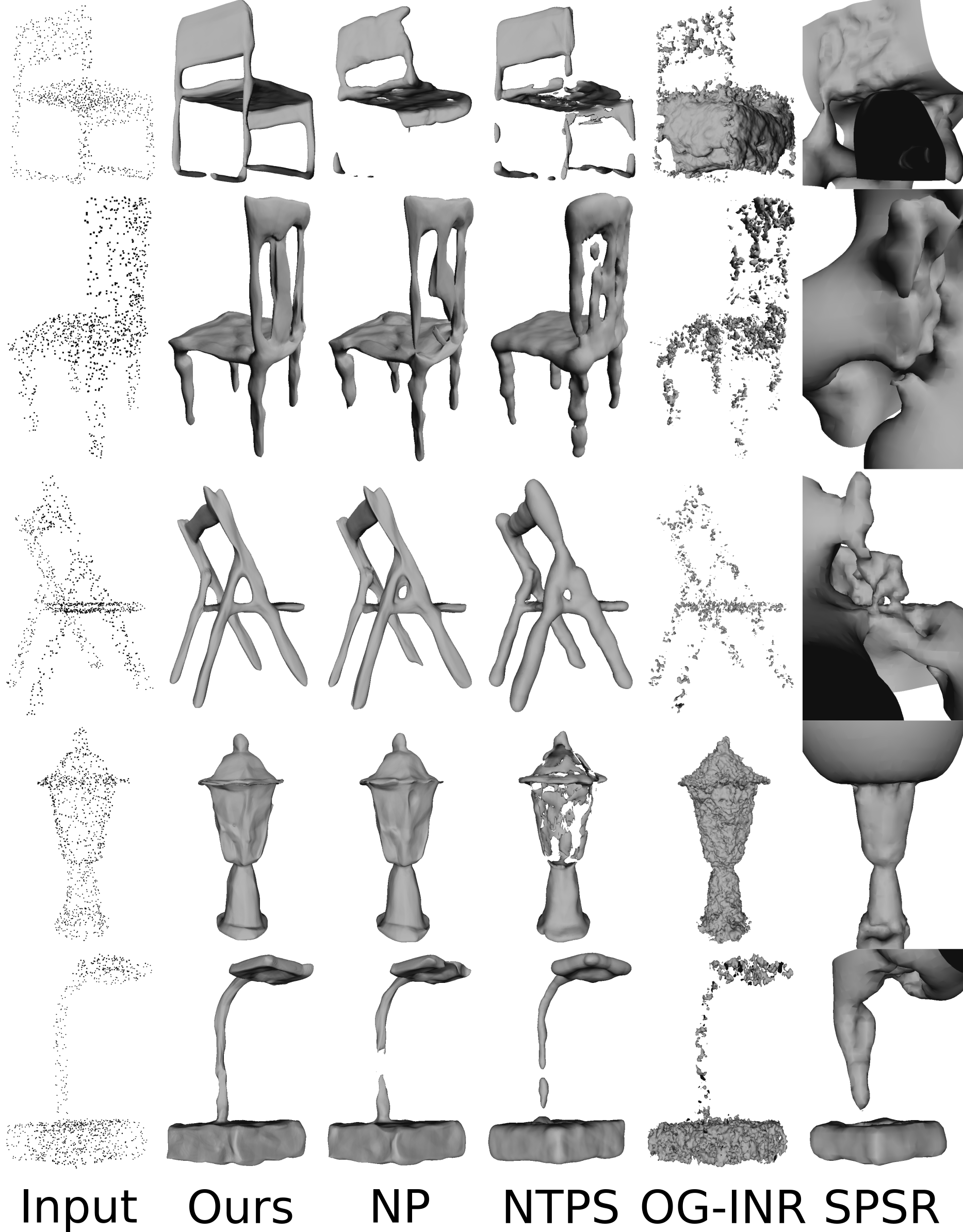}
\caption{ShapeNet \cite{shapenet} reconstructions from sparse noisy unoriented point clouds.}
\label{fig:sn}
\end{figure}

\begin{table}[h!]
\centering
\scalebox{0.82}{
\begin{tabular}{lllll}
\hline
       & \textbf{CD1}     & \textbf{CD2}    & \textbf{NC}       & \textbf{FS}       \\ \hline
SPSR \cite{kazhdan2013screened}     & 2.34          & 0.224          & 0.74          & 0.50         \\
OG-INR \cite{koneputugodage2023octree}  & 1.36          & 0.051          & 0.55          & 0.55          \\
NP \cite{ma2020neural}     & 1.16          & 0.074          & 0.84          & 0.75          \\
GP \cite{chen2023gridpull}  &1.07	&0.032	& 0.70 & 0.74\\
NTPS \cite{NeuralTPS}    & 1.11          & 0.067          & \textbf{0.88} & 0.74         \\
Ours   & \textbf{0.76} & \textbf{0.020} & 0.87         & \textbf{0.83} \\ \hline
\end{tabular}}
%\vspace{-5pt}
\caption{ShapeNet \cite{shapenet} reconstructions from sparse noisy unoriented point clouds.}
\label{tab:sn}
\end{table}

\begin{table}[h!]
\centering
\scalebox{0.85}{
\begin{tabular}{lllll}
\hline
               & \textbf{CD1}     & \textbf{CD2} & \textbf{NC}       & \textbf{FS} \\ \hline

POCO \cite{boulch2022poco}  & 0.308     & 0.002   & 0.934        & 0.981    \\
CONet \cite{peng2020convolutional}  & 1.260     & 0.048    & 0.829      & 0.599    \\
On-Surf \cite{ma2022reconstructing} &0.584    &0.012   &0.936 & 0.915  \\
SAC \cite{tang2021sa}  &0.261    &0.002   &0.935 & 0.975  \\
NKSR \cite{Huang_2023_CVPR} & 0.274    &0.002   &0.945 & 0.981  \\
\hline
SPSR \cite{kazhdan2013screened}    & 0.751     & 0.028    & 0.871        & 0.839    \\
GP \cite{chen2023gridpull}  &0.495    & 0.005   &0.887 & 0.945\\
NTPS \cite{NeuralTPS}  & 0.737     & 0.015     & 0.943          & 0.844     \\
NP \cite{ma2020neural}  & 0.269    & 0.003     & 0.951          & 0.973   \\
Ours  & \textbf{0.220}    & \textbf{0.001}     & \textbf{0.956} & \textbf{0.981}    \\ \hline
\end{tabular}}
\label{tab:fs}
%\vspace{-5pt}
\caption{Faust \cite{Bogo:CVPR:2014} reconstructions from sparse noisy unoriented point clouds.}
\label{tab:fs}
\end{table}

\begin{figure}[t!]
\centering
\includegraphics[width=\linewidth]{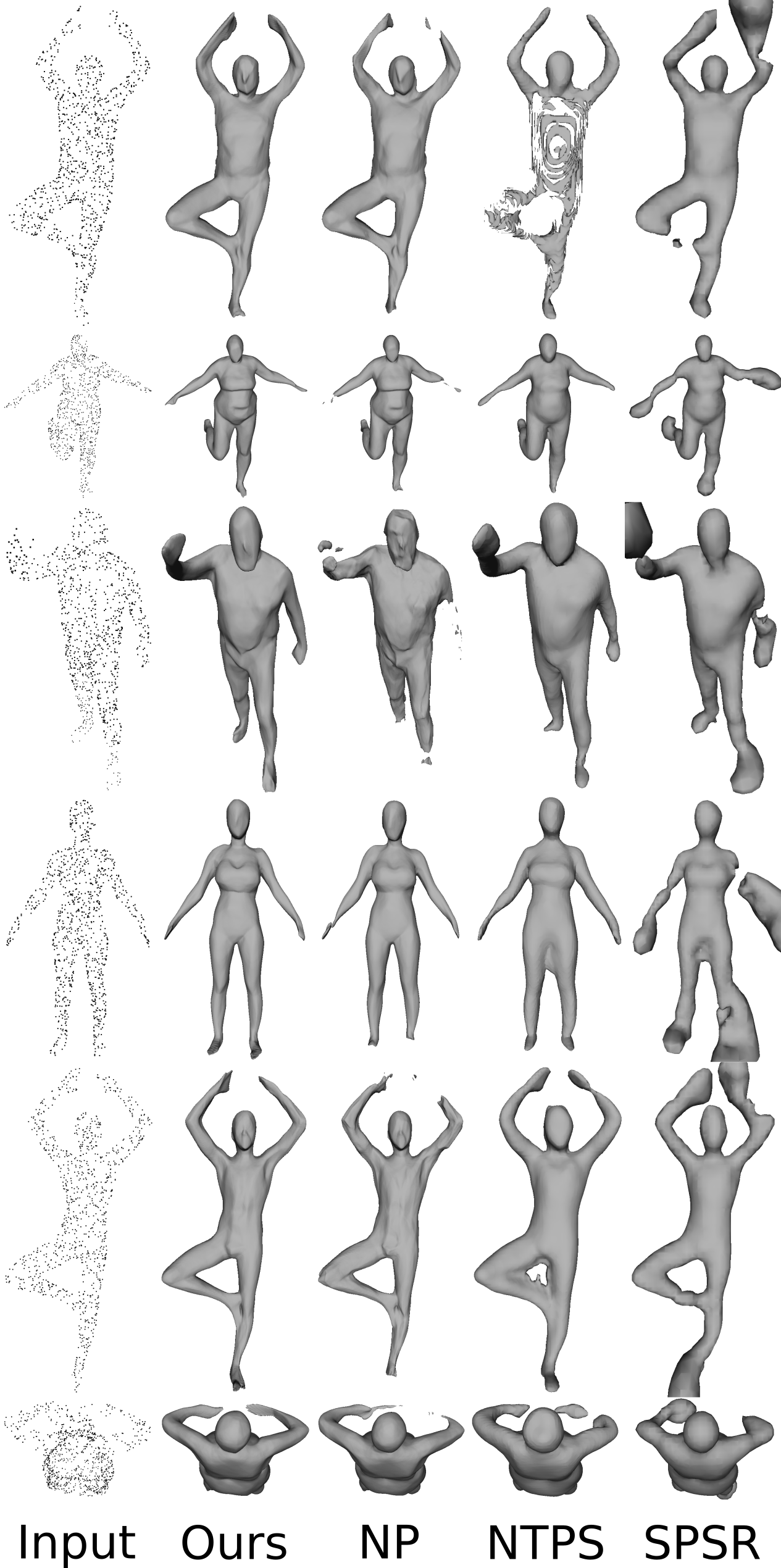}
\caption{Faust \cite{Bogo:CVPR:2014} reconstructions from sparse noisy unoriented point clouds.}
\label{fig:fs}
\end{figure}

\begin{table*}[t!]
\centering
\scalebox{0.65}{
\begin{tabular}{l|lllllllllllllllllll}
\cline{1-19}
\multicolumn{1}{c|}{} & \multicolumn{3}{c}{\textbf{Burghers}}                                                       & \multicolumn{3}{c}{\textbf{Copyroom}}                                                       & \multicolumn{3}{c}{\textbf{Lounge}}                                                         & \multicolumn{3}{c}{\textbf{Stonewall}}                                                      & \multicolumn{3}{c}{\textbf{Totemple}}                                                       & \multicolumn{3}{c}{\textbf{Mean}}                                    &  \\ \cline{2-19}
\multicolumn{1}{c|}{}                  & \multicolumn{1}{l|}{CD1}  & \multicolumn{1}{l|}{CD2}  & \multicolumn{1}{l|}{NC}    & \multicolumn{1}{l|}{CD1}  & \multicolumn{1}{l|}{CD2}  & \multicolumn{1}{l|}{NC}    & \multicolumn{1}{l|}{CD1}  & \multicolumn{1}{l|}{CD2}  & \multicolumn{1}{l|}{NC}    & \multicolumn{1}{l|}{CD1}  & \multicolumn{1}{l|}{CD2}  & \multicolumn{1}{l|}{NC}    & \multicolumn{1}{l|}{CD1}  & \multicolumn{1}{l|}{CD2}  & \multicolumn{1}{l|}{NC}    & \multicolumn{1}{l|}{CD1} & \multicolumn{1}{l|}{CD2} & NC    &  \\ \cline{1-19}
SPSR \cite{kazhdan2013screened}                                  & 0.178                     & 0.205                     & \multicolumn{1}{l|}{0.874} & 0.225                     & 0.286                     & \multicolumn{1}{l|}{0.861} & 0.280                     & 0.365                     & \multicolumn{1}{l|}{0.869} & 0.300                     & 0.480                     & \multicolumn{1}{l|}{0.866} & 0.588                     & 1.673                     & \multicolumn{1}{l|}{0.879} & 0.314                    & 0.602                    & 0.870 &  \\
NDrop \cite{boulch2021needrop}                                 & 0.200                     & 0.114                     & \multicolumn{1}{l|}{0.825} & 0.168                     & 0.063                     & \multicolumn{1}{l|}{0.696} & 0.156                     & 0.050                     & \multicolumn{1}{l|}{0.663} & 0.150                     & 0.081                     & \multicolumn{1}{l|}{0.815} & 0.203                     & 0.139                     & \multicolumn{1}{l|}{0.844} & 0.175                    & 0.089                    & 0.769 &  \\
NP \cite{ma2020neural}                                    & 0.064                     & 0.008                     & \multicolumn{1}{l|}{0.898} & 0.049                     & 0.005                     & \multicolumn{1}{l|}{0.828} & 0.133                     & 0.038                     & \multicolumn{1}{l|}{0.847} & 0.060                     & 0.005                     & \multicolumn{1}{l|}{0.910} & 0.178                     & 0.024                     & \multicolumn{1}{l|}{0.908} & 0.097                    & 0.016                    & 0.878 &  \\
SAP \cite{peng2021shape}                                    & 0.153                     & 0.101                     & \multicolumn{1}{l|}{0.807} & 0.053                     & 0.009                     & \multicolumn{1}{l|}{0.771} & 0.134                     & 0.033                     & \multicolumn{1}{l|}{0.813} & 0.070                     & 0.007                     & \multicolumn{1}{l|}{0.867} & 0.474                     & 0.382                     & \multicolumn{1}{l|}{0.725} & 0.151                    & 0.100                    & 0.797 &  \\
NSpline \cite{williams2021neural}                              & 0.135                     & 0.123                     & \multicolumn{1}{l|}{0.891} & 0.056                     & 0.023                     & \multicolumn{1}{l|}{0.855} & 0.063                     & 0.039                     & \multicolumn{1}{l|}{0.827} & 0.124                     & 0.091                     & \multicolumn{1}{l|}{0.897} & 0.378                     & 0.768                     & \multicolumn{1}{l|}{0.892} & 0.151                    & 0.209                    & 0.88  &  \\
NTPS \cite{NeuralTPS}                                  & 0.055                     & 0.005                     & \multicolumn{1}{l|}{\textbf{0.909}} & 0.045                     & 0.003                     & \multicolumn{1}{l|}{\textbf{0.892}} & 0.129                     & 0.022                     & \multicolumn{1}{l|}{\textbf{0.872}} & 0.054                     & 0.004                     & \multicolumn{1}{l|}{\textbf{0.939}} & 0.103                     & 0.017                     & \multicolumn{1}{l|}{0.935} & 0.077                    & 0.010                    & \textbf{0.897} &  \\ \cline{1-19}
Ours                                   & \multicolumn{1}{r}{\textbf{0.051}} & \multicolumn{1}{r}{\textbf{0.006}} & \multicolumn{1}{r|}{0.881}  & \multicolumn{1}{r}{\textbf{0.037}} & \multicolumn{1}{r}{\textbf{0.002}} & \multicolumn{1}{r|}{0.833}  & \multicolumn{1}{r}{\textbf{0.044}} & \multicolumn{1}{r}{\textbf{0.011}} & \multicolumn{1}{r|}{0.862}  & \multicolumn{1}{r}{\textbf{0.035}} & \multicolumn{1}{r}{\textbf{0.003}} & \multicolumn{1}{r|}{0.912}  & \multicolumn{1}{r}{\textbf{0.042}} & \multicolumn{1}{r}{\textbf{0.002}} & \multicolumn{1}{r|}{0.925} & \textbf{0.041}                    & \textbf{0.004}                    & 0.881 &  \\ \cline{1-19}
\end{tabular}
}
%\vspace{-5pt}
\caption{3D Scene (\cite{zhou2013dense}) reconstructions from sparse  point clouds.}
\label{tab:3ds}
\end{table*}

\subsection{Object level reconstruction}
We perform reconstruction of ShapeNet \cite{shapenet} objects from sparse and noisy point clouds. Table \ref{tab:sn} and Figure \ref{fig:sn} show respectively a numerical and qualitative comparison to the competition. 
We outperform the competition across all metrics, as witnessed by the visual superiority of our reconstructions. We manage to recover fine structures and details with more fidelity. Although it obtains overall good coarse reconstructions, the thin plate spline smoothing prior of NTPS seems to be hindering its expressivity. We noticed OG-INR fails to converge to satisfactory results under the sparse and noisy regime despite its effective Octree based sign field guidance in dense settings.

\subsection{Real articulated shape reconstruction}

We perform reconstruction of Faust (\cite{shapenet}) human shapes from sparse and noisy point clouds. Table \ref{tab:fs} and Figure \ref{fig:fs} show respectively a numerical and qualitative comparison to the competition. We outperform the other methods across all metrics. Visually, our reconstructions are particularly better at body extremities. Similarly to fine structures in the ShapeNet experiment, these are areas where the input point cloud samples are scarce and shape prediction is hard and ambiguous. NTPS reconstructions tend to be coarser and less detailed on this data as well. Notice that ShapeNet trained Generalizable methods (Top half of the table) do not all necessarily generalize well in this experiment. 

\begin{figure}[t!]
\centering
\includegraphics[width=1.0\linewidth]{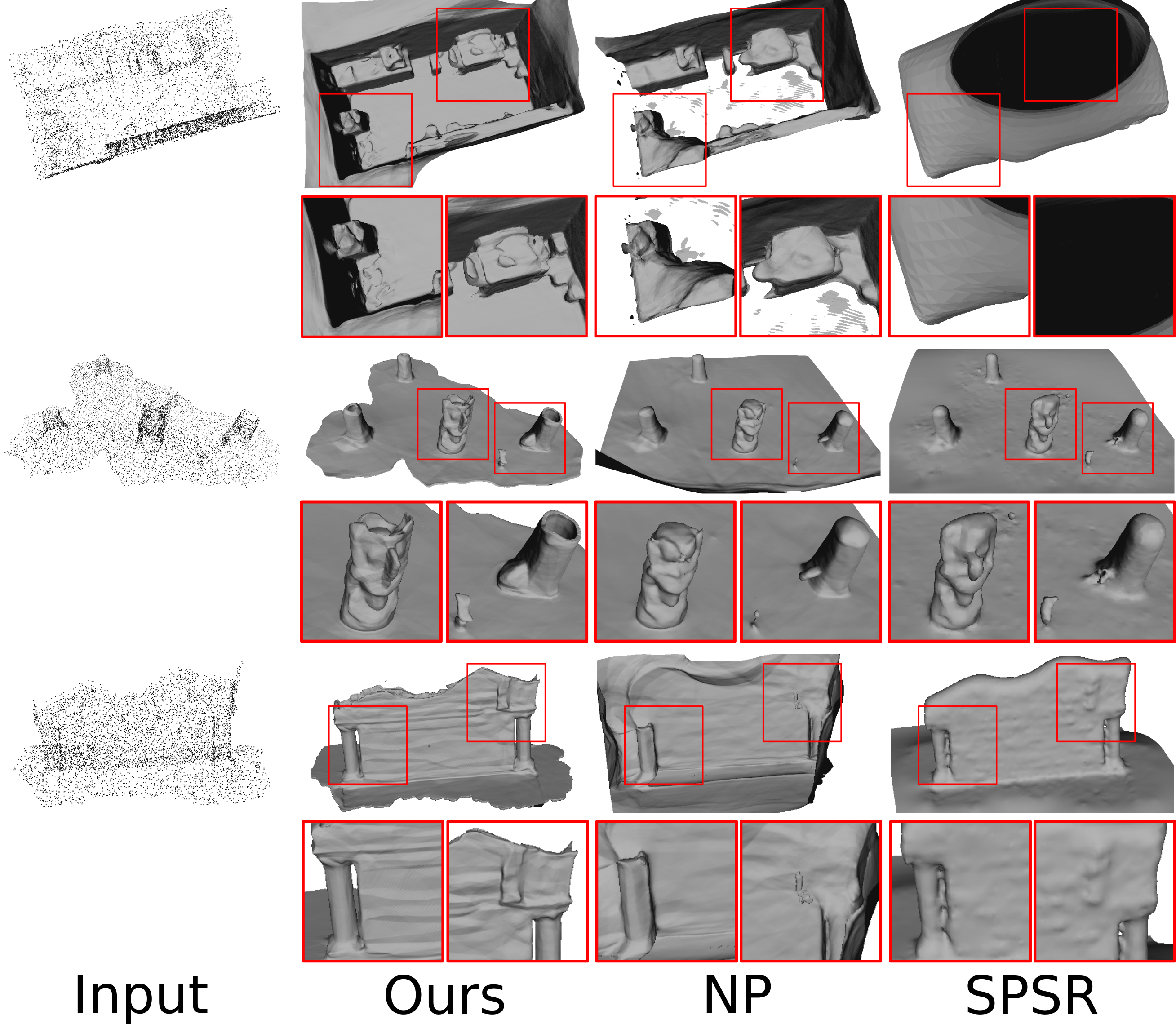}
%\vspace{-10pt}
\caption{3D Scene \cite{zhou2013dense} reconstructions from sparse unoriented point clouds.}
\label{fig:3ds}
\end{figure}

\subsection{Real scene level reconstruction}

Following \cite{NeuralTPS}, we report reconstruction results on the 3D Scene (\cite{zhou2013dense}) data from spatially sparse point clouds. Table \ref{tab:3ds} summarizes numerical results. We compiled results for methods NTPS, NP, SAP, NDrop and NSpline as reported in state-of-the-art method NTPS.  We outperform the competition is this benchmark thanks to our loss, as our baseline NP displays more blatant failures in this large scale sparse setup. Figure \ref{fig:3ds} shows qualitative comparisons to our baseline NP and SPSR. Red boxes highlight areas where our method displays particularly better details and fidelity in the reconstruction. 

\subsection{Varying the point cloud density}

\begin{figure*}[t!]
\centering
\includegraphics[width=.7\linewidth]{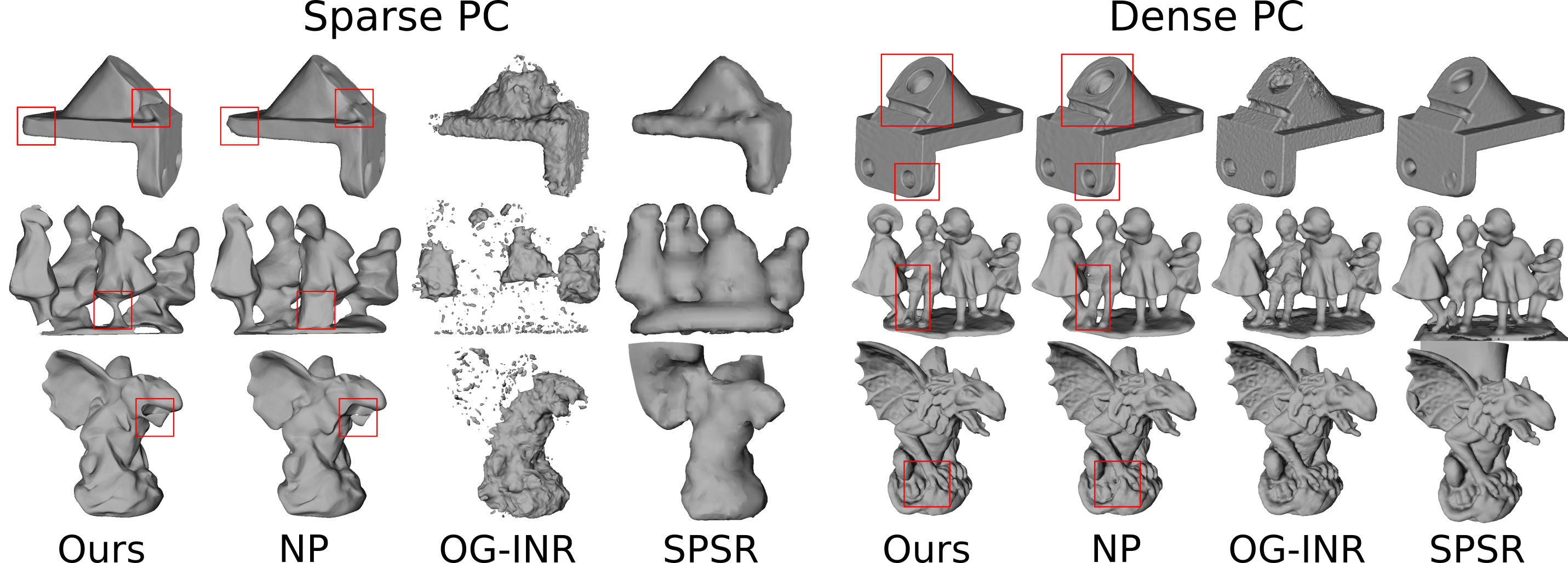}
%\vspace{-10pt}
\caption{SRB \cite{williams2019deep} reconstructions from unoriented sparse and dense inputs.}
\label{fig:srb}
\end{figure*}

We use the SRB \cite{williams2019deep} benchmark to assess the behavior of our method across different point cloud densities. Table \ref{tab:srb} indicates comparative results under both $1024$ sized and dense input point clouds. We compiled results for the competition from OG-INR in the dense setting. We outperform our competition in the sparse case, and we perform on par with the state-of-the-art in the dense case. Our improvement \wrt our baseline (NP) is substantial for both sparse and dense inputs. This can be seen visually in Figure \ref{fig:srb}, where we show reconstructions for both sparse and dense cases. Notice how we recover better topologies in the sparse case and improved and more accurate details in the dense case, as pinpointed by the red boxes.  
These results showcase the utility and benefit of our contribution even in the dense setting. We note that SPSR becomes a very viable contestant qualitatively in the dense setting.

\section{Ablation studies}

The ablation of our main contribution is present throughout all  Tables and Figures. In fact while we use the combined loss in Equation \ref{equ:final}, our baseline (\ie NP) uses solely the query projection loss in Equation \ref{equ:np}. The improvement brought by our additional loss is visible across real/synthetic, scenes/objects, sparse/dense point clouds.

\begin{table}[t!]
\centering
\scalebox{0.90}{
\begin{tabular}{lcc}
\hline
& Sparse & Dense \\
\hline
SPSR \cite{kazhdan2013screened}   & 2.27 & 1.25 \\
DIGS \cite{ben2022digs}  & 0.68 & \textbf{0.19}\\
OG-INR \cite{koneputugodage2023octree}  & 0.85 &  0.20\\
NTPS \cite{NeuralTPS}  & 0.73 & - \\
NP \cite{ma2020neural}  & 0.58 & 0.23  \\
Ours  & \textbf{0.49} & \textbf{0.19} \\ \hline
\end{tabular}}
\caption{Ablation of point cloud density}
\label{tab:srb}
\end{table}

\begin{table}[h!]
\centering
\begin{tabular}[t]{lll}
\hline
%\firsthline
\textbf{}                & \textbf{CD1}             & \textbf{NC}  \\ \hline
NP \cite{ma2020neural} (baseline)                   & 1.10          & 0.85 \\ %\hline
Ours (local) $\rho_q$  = $\sigma_p$/10    & 0.92          & 0.86 \\
Ours (local) $\rho_q$ = $\sigma_p$/100   & \textbf{0.75} & 0.86 \\
Ours (local) $\rho_q$ = $\sigma_p$/500   & 0.88          & 0.86 \\
Ours (local) $\rho_q$ = $\sigma_p$/1000  & 1.02          & 0.84 \\
Ours (global ) $\rho$ = $\sigma$/100 & 0.98          & 0.86 \\ \hline
\end{tabular}
 %\vspace{-5pt}
\caption{Ablation of hard sample search radii.}
\label{tab:abl}
\end{table}

\textbf{Perturbation radii} 
 We perform an ablation of using local \vs global radii $\rho$ (Equation \ref{equ:delta}) and the choice of value of local radii $\rho_q$ in Table \ref{tab:abl}. Results show that using local radii ($\rho_q$) is a superior strategy as it intuitively allows for a spatially adaptive search of hard samples (global radius is $\sigma\times 10^{-2}$, where $\sigma$ is the average $\sigma_p$). We note that our baseline NP constitutes the special case $\rho_q=0$. A value of $\sigma_p \times 10^{-2}$ achieves empirically satisfactory results 
($p$ being the nearest point to the query $q$ in the input point cloud). 
Decreasing $\rho_q$ leads expectedly to worse results as less hard queries are available for sampling.  However, we also note that very large values of $\rho_q$ can lead to spurious pseudo supervision, as adversarial samples $q+\delta$ run the risk of no longer having the same nearest point in the input point cloud as their original sample $q$. We performed this ablation on class Table of ShapeNet.

\textbf{Multitask loss}
To guarantee the stability of our learning process, we employ a hybrid loss that combines the original objective with an adversarial one. This approach becomes crucial when a shape-specific trade-off between adversarial regularization and the original loss is required for the convergence of the shape function. In practical terms, this strategy outperformed using the adversarial loss alone, leading to an improvement in CD1 from 0.78 to 0.75 in class Table of ShapeNet.

\textbf{Increasing the number of query points}
We increase the number of NP original query samples to equal the total number of samples in our method (\ie original queries + adversarial queries). We find that the performance of NP with extra queries only leads occasionally to marginal improvement (On average Chamfer distance goes from 0.581 to 0.576 in SRB). 

\begin{figure}[t!]
\centering
\includegraphics[width=.6\linewidth]{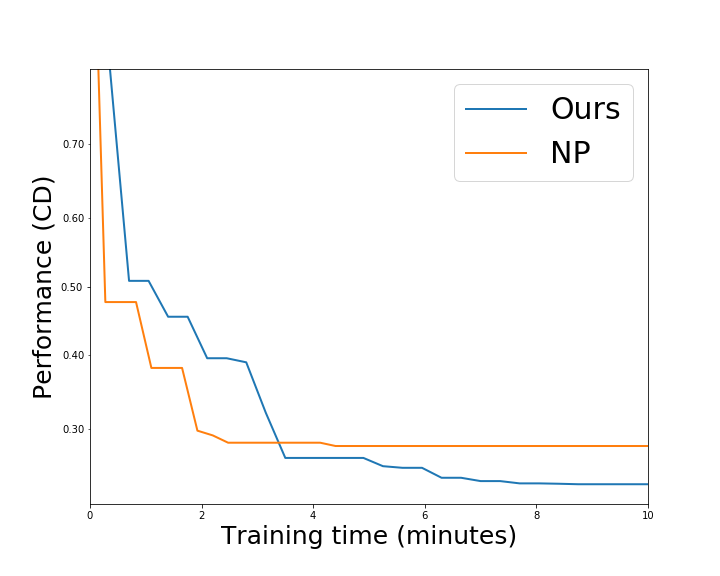}
\caption{Training time for shape Gargoyle of benchmark SRB \cite{williams2019deep}.}
\label{fig:time}
\end{figure}

\textbf{Training time}
We provide in Figure \ref{fig:time} a plot showing the best performance attained by our method and the baseline (NP) \wrt the training time (RTX A6000). As it can be seen in this figure, within less than 4 minutes of training, we already surpass the best baseline performance. NP's performance stagnates from that point onwards due to overfitting, while our performance keeps on improving. We reach our optimal result after 8 minutes of training. The experiment was conducted on the challenging shape Gargoyle of benchmark SRB using a 1024 sized input point cloud.

\section{Limitations}

 As can be seen in \eg Figure \ref{fig:sn}, even though we improve on our baseline, we still face difficulties in learning very thin and convoluted shape details, such as chair legs and slats. Although our few-shot problem is inherently challenging and can be under-constrained in many cases, we still believe there is room for improvement for our proposed strategy in this setting. For instance, one element that could be a source of hindrance to our accuracy is the difficulty of balancing empirical risk minimization and adversarial learning. In this work, we used an off-the-shelf self-trained loss weighting strategy, and we would like to address this challenge further as part of our future work.

\section{Conclusion}

We explored in this work a novel idea for regularizing implicit shape representation learning from sparse unoriented point clouds. We showed that harnessing adversarial samples locally in training can lead to numerous desirable outcomes, including superior results, reduced over fitting and easier evaluation model selection.

\section*{Impact statement}
This paper presents work whose goal is to advance the fields of Machine Learning and 3D Computer Vision, specifically implicit neural shape representation learning. There are many potential societal consequences of our work, none which we feel must be specifically highlighted here.

%\bibliography{example_paper}
\bibliography{main}
\bibliographystyle{icml2024}

%%%%%%%%%%%%%%%%%%%%%%%%%%%%%%%%%%%%%%%%%%%%%%%%%%%%%%%%%%%%%%%%%%%%%%%%%%%%%%%
%%%%%%%%%%%%%%%%%%%%%%%%%%%%%%%%%%%%%%%%%%%%%%%%%%%%%%%%%%%%%%%%%%%%%%%%%%%%%%%
% APPENDIX
%%%%%%%%%%%%%%%%%%%%%%%%%%%%%%%%%%%%%%%%%%%%%%%%%%%%%%%%%%%%%%%%%%%%%%%%%%%%%%%
%%%%%%%%%%%%%%%%%%%%%%%%%%%%%%%%%%%%%%%%%%%%%%%%%%%%%%%%%%%%%%%%%%%%%%%%%%%%%%%
\newpage
\appendix
\onecolumn

\section{Additional visualizations}

\begin{figure}[h!]
\centering
\includegraphics[width=.7\linewidth]{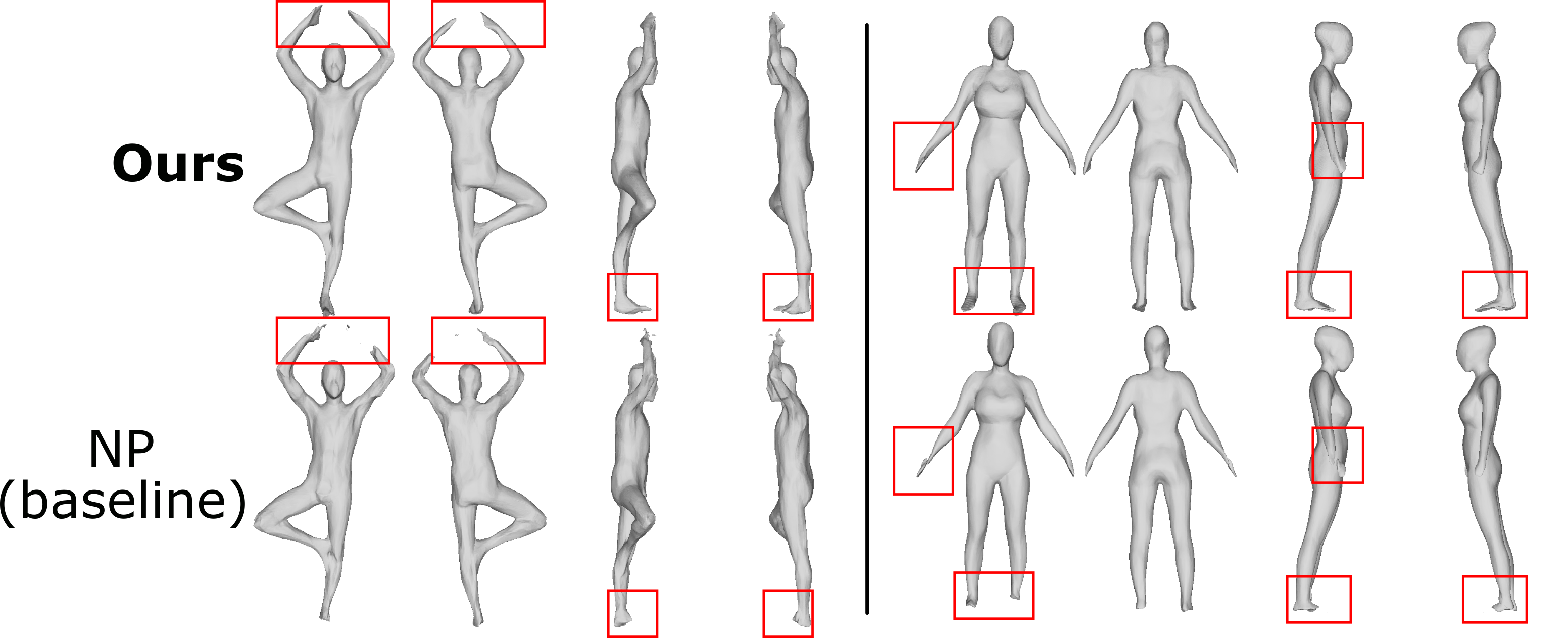}
\caption{Faust \cite{Bogo:CVPR:2014} reconstructions from sparse noisy unoriented point clouds (1024 pts).}
\label{fig:fsx}
\end{figure}

\begin{figure}[h!]
\centering
\includegraphics[width=.7\linewidth]{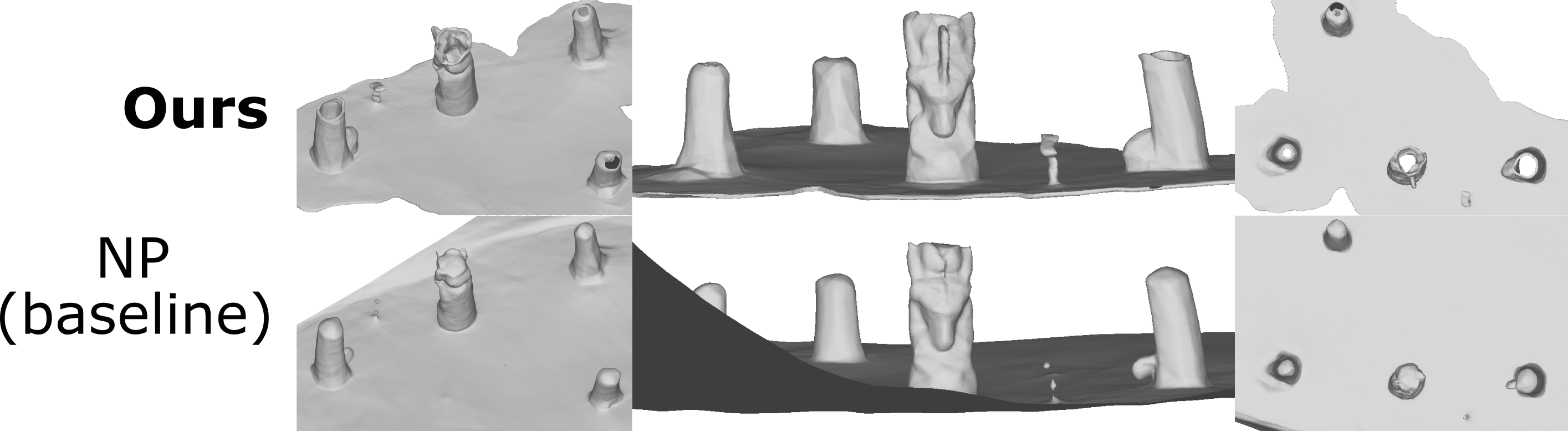}
\caption{3D Scene \cite{zhou2013dense} reconstructions from sparse unoriented point clouds (100 pts per m$^2$).}
\label{fig:totx}
\end{figure}

\begin{figure}[h!]
\centering
\includegraphics[width=.7\linewidth]{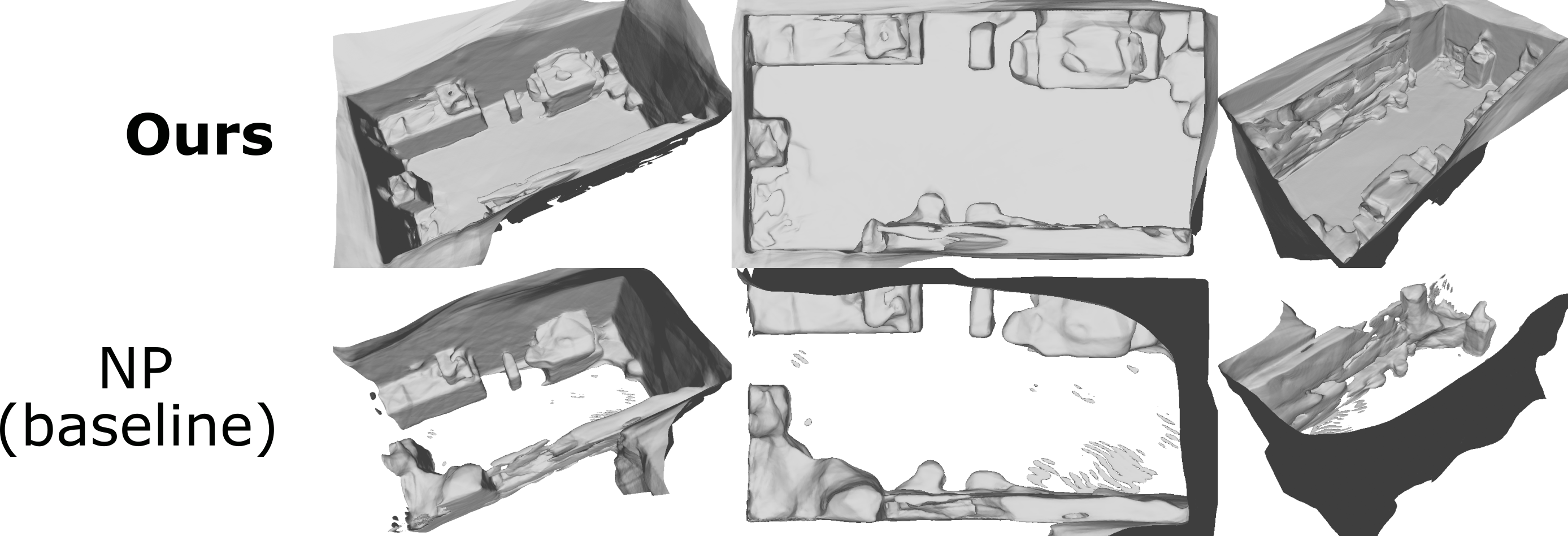}
\caption{3D Scene \cite{zhou2013dense} reconstructions from sparse unoriented point clouds (100 pts per m$^2$).}
\label{fig:copx}
\end{figure}

Figures \ref{fig:fsx},\ref{fig:totx},\ref{fig:copx} show more multi-view qualitative comparisons to our baseline NP. 

\section{Metrics}

Following the definitions from \cite{boulch2022poco} and \cite{williams2019deep}, we present here the formal definitions for the metrics that we use for evaluation in the main submission. We denote by $\mathcal{S}$ and  $\hat{\mathcal{S}}$ the ground truth and predicted mesh respectively. All metrics are approximated with 100k samples from the groundtruth mesh $\mathcal{S}$ and reconstruction $\hat{\mathcal{S}}$. 

\paragraph{Chamfer Distance (CD$_1$)} The L$_1$ Chamfer distance is based on the two-ways nearest neighbor distance: 
$$\mathrm{CD}_1=\frac{1}{2|\mathcal{S}|} \sum_{v \in \mathcal{S}} \min _{\hat{v} \in \hat{\mathcal{S}}}\|v-\hat{v}\|_2+\frac{1}{2|\hat{\mathcal{S}}|} \sum_{\hat{v} \in \hat{\mathcal{S}}} \min _{v \in \mathcal{S}}\|\hat{v}-v\|_2.$$

\paragraph{Chamfer Distance (CD$_2$)} The L$_2$ Chamfer distance is based on the two-ways nearest neighbor squared distance: 
$$\mathrm{CD}_2=\frac{1}{2|\mathcal{S}|} \sum_{v \in \mathcal{S}} \min _{\hat{v} \in \hat{\mathcal{S}}}\|v-\hat{v}\|_2^2+\frac{1}{2|\hat{\mathcal{S}}|} \sum_{\hat{v} \in \hat{\mathcal{S}}} \min _{v \in \mathcal{S}}\|\hat{v}-v\|_2^2.$$

\paragraph{F-Score (FS)} For a given threshold $\tau$, the F-score between  the meshes $\mathcal{S}$ and $\hat{\mathcal{S}}$ is defined as:
$$
\mathrm{FS}\left(\tau, \mathcal{S}, \hat{\mathcal{S}}\right)=\frac{2 \text { Recall} \cdot \text{Precision }}{\text { Recall }+\text { Precision }},
$$

where
$$
\begin{array}{r}
\operatorname{Recall}\left(\tau, \mathcal{S}, \hat{\mathcal{S}}\right)=\mid\left\{v \in \mathcal{S} \text {, s.t. } \min _{\hat{v} \in \hat{ \mathcal{S} }} \left\|v-\hat{v}\|_2\right<\tau\right\} \mid ,\\
\operatorname{Precision}\left(\tau, \mathcal{S}, \hat{\mathcal{S}}\right)=\mid\left\{\hat{v} \in \hat{\mathcal{S} }\text {, s.t. } \min _{v \in  \mathcal{S} } \left\|v-\hat{v}\|_2\right<\tau\right\} \mid .\\
\end{array}
$$
Following \cite{mescheder2019occupancy} and \cite{peng2020convolutional}, we set $\tau$ to $0.01$.

\paragraph{Normal consistency (NC)} We denote here by $n_v$ the normal at a point $v$ in $\mathcal{S}$. The normal consistency between two meshes $\mathcal{S}$ and $\hat{\mathcal{S}}$ is defined as: 

$$\mathrm{NC}=\frac{1}{2|\mathcal{S}|} \sum_{v \in \mathcal{S}} n_{v} \cdot n_{\operatorname{closest}(v,\hat{ \mathcal{S}})}+\frac{1}{2|\hat{\mathcal{S}}|} \sum_{\hat{v} \in \hat{\mathcal{S}}} n_{\hat{v}} \cdot n_{\operatorname{closest}(\hat{v}, \mathcal{S})},$$

where  
$$
\operatorname{closest}(v, \hat{\mathcal{S}}) = \operatorname{argmin} _{\hat{v} \in \hat{\mathcal{S}}}\|v-\hat{v}\|_2.
$$

%%%%%%%%%%%%%%%%%%%%%%%%%%%%%%%%%%%%%%%%%%%%%%%%%%%%%%%%%%%%%%%%%%%%%%%%%%%%%%%
%%%%%%%%%%%%%%%%%%%%%%%%%%%%%%%%%%%%%%%%%%%%%%%%%%%%%%%%%%%%%%%%%%%%%%%%%%%%%%%

\end{document}